\pdfoutput=1

\documentclass[11pt]{article}

\usepackage[]{EACL2023}

\usepackage{times}
\usepackage{latexsym}

\usepackage[T1]{fontenc}

\usepackage[utf8]{inputenc}

\usepackage{microtype}

\usepackage{inconsolata}

\usepackage{booktabs}
\usepackage{amsmath,amssymb}
\usepackage{pifont}
\usepackage{xcolor}
\usepackage{listings}
\usepackage[scaled]{beramono}

\usepackage{makecell}
\usepackage{siunitx, mhchem}
\usepackage{graphicx}
\usepackage[export]{adjustbox}

\usepackage{xspace}
\newcommand{\placeholder}{\textsl{ferret}\xspace} 
\newcommand{\greencheck}{{\color{teal}\ding{51}}}
\newcommand{\redcross}{{\color{purple}\ding{55}}}

\newcommand{\repo}{\url{https://github.com/g8a9/ferret}\xspace}
\newcommand{\urldemo}{\url{https://huggingface.co/spaces/g8a9/ferret}\xspace}
\newcommand{\urlvideo}{\url{https://youtu.be/kX0HcSah_M4}\xspace}

\title{\placeholder: a Framework for Benchmarking Explainers on Transformers}

\author{Giuseppe Attanasio$^{\clubsuit}$, Eliana Pastor$^{\diamondsuit}$, Chiara Di Bonaventura$^{\spadesuit}$, Debora Nozza$^{\clubsuit}$ \\ \\
 $^{\clubsuit}$Bocconi University, Milan, Italy \\
 $^{\diamondsuit}$Politecnico di Torino, Turin, Italy\\
 $^{\spadesuit}$King's College London, London, United Kingdom 
 \\
 { \tt \{giuseppe.attanasio3,debora.nozza\}@unibocconi.it} \\
 { \tt eliana.pastor@polito.it}  \\
 \texttt{chiara.di\_bonaventura@kcl.ac.uk} }

\begin{document}
\maketitle
\begin{abstract}
As Transformers are increasingly relied upon to solve complex NLP problems, there is an increased need for their decisions to be humanly interpretable. 
While several explainable AI (XAI) techniques for interpreting the outputs of transformer-based models have been proposed, there is still a lack of easy access to using and comparing them.
We introduce \placeholder, a Python library to simplify the use and comparisons of XAI methods on transformer-based classifiers.
With \placeholder, users can visualize and compare transformers-based models output explanations using state-of-the-art XAI methods on any free-text or existing XAI corpora. Moreover, users can also evaluate ad-hoc XAI metrics to select the most faithful and plausible explanations. 
To align with the recently consolidated process of sharing and using transformers-based models from Hugging Face, \placeholder interfaces directly with its Python library.
In this paper, we showcase \placeholder to benchmark XAI methods used on transformers for sentiment analysis and hate speech detection. We show how specific methods provide consistently better explanations and are preferable in the context of transformer models.

\end{abstract}

\section{Introduction}

Transformers have revolutionized NLP applications in recent years due to their strong performance on various tasks; their black-box nature remains an obstacle for practitioners who need explanations about why specific predictions were made and what features drove them. The development of explainable AI (XAI) techniques on several NLP tasks \cite{10.1145/3546577} has helped bridge this gap by providing insight into the inner workings of transformers and helping users gain trust in their decisions.
Several XAI approaches have been proposed in the literature \cite{ribeiro2016should,NIPS2017_lundberg_SHAP,DBLP:journals/corr/SimonyanVZ13, pastor-lace-2019}, also tailored to Transformer models \cite{wallace-etal-2019-universal,DBLP:journals/corr/LiMJ16a,jin2019towards,ross-etal-2021-explaining}.
Despite the importance of making XAI methods accessible to NLP experts and practitioners through practical tools, there is still a lack of accessibility for transformer models. 
XAI for transformers is mainly scattered and hard to operationalize. Methods come with independent implementations or framework-specific libraries that do not allow either evaluation or cross-method comparison. Further, existing implementations are not integrated with widespread transformers libraries (e.g., Hugging Face's \textit{transformers} \cite{wolf-etal-2020-transformers}).
The lack of standardization and weak interoperability leaves practitioners with unsolved questions, such as choosing the \textit{best} method given a task and a model \cite{attanasio-etal-2022-benchmarking}.

We introduce \placeholder (FramEwork foR benchmaRking Explainers on Transformers), an open-source Python library that drastically simplifies the use and comparison of XAI methods on transformers.
The library stems from vertical scientific contributions and focused engineering efforts.
On the one hand,
\placeholder provides the first-of-its-kind API (see Figure~\ref{fig:example_code}) to use and compare explanation methods along the established criteria of faithfulness and plausibility \cite{jacovi-goldberg-2020-towards}.
On the other hand,
it integrates seamlessly with \textit{transformers} \cite{wolf-etal-2020-transformers}, making it an easy add-on to existing Transformer-based pipelines and NLP tasks.
\placeholder permits to run four state-of-the-art XAI methods, compute six ad-hoc XAI evaluation metrics, and easily load four existing interpretability datasets. Further, it offers abstract interfaces to foster future integration of methods, metrics, and datasets. 

We showcase \placeholder on sentiment analysis and hate speech detection case studies. Faithfulness and plausibility metrics highlight SHAP \cite{NIPS2017_lundberg_SHAP} as the most consistent explainer on single- and multiple-samples scenarios.



\paragraph{Contributions.} We release \placeholder, the first-of-its-kind benchmarking framework for interpretability tightly integrated with Hugging Face's \textit{transformers} library.
We release our code and documentation,\footnote{\repo} an interactive demo,\footnote{\urldemo} and a video tutorial.\footnote{\urlvideo}


\definecolor{codegreen}{rgb}{0,0.6,0}
\definecolor{codegray}{rgb}{0.5,0.5,0.5}

\definecolor{backcolour}{RGB}{245,248,250}
\definecolor{emph}{RGB}{166,88,53}
\definecolor{nightblue}{RGB}{9,49,105}
\definecolor{keywords}{RGB}{207,33,46}
\definecolor{lightpurple}{RGB}{130,81,223}

\lstdefinestyle{mystyle}{
    backgroundcolor=\color{backcolour},   
    commentstyle=\color{codegreen},
    keywordstyle=\color{keywords},
    stringstyle=\color{nightblue},
    basicstyle=\ttfamily\footnotesize,
    breakatwhitespace=false,         
    breaklines=true,                 
    captionpos=b,                    
    keepspaces=true,                 
    showspaces=false,                
    showstringspaces=false,
    showtabs=false,                  
    tabsize=2,
    frame=shadowbox,
    emph={AutoTokenizer,AutoModelForSequenceClassification,Explainer},
    emphstyle={\color{emph}},
    emph={[2]from_pretrained,compute_table},
    emphstyle={[2]\color{lightpurple}}
}

\lstset{style=mystyle}

\begin{figure*}[!t]
    \centering
    \begin{lstlisting}[language=Python]
from transformers import AutoModelForSequenceClassification, AutoTokenizer
from ferret import Benchmark

name = "cardiffnlp/twitter-xlm-roberta-base-sentiment"
model = AutoModelForSequenceClassification.from_pretrained(name)
tokenizer = AutoTokenizer.from_pretrained(name)

bench = Benchmark(model, tokenizer)
explanations = bench.explain("You look stunning!", target=1)
evaluations = bench.evaluate_explanations(explanations, target=1)
\end{lstlisting}
    \caption{Essential code to benchmark explanations on an existing Hugging Face model using \placeholder.}
    \label{fig:example_code}
\end{figure*}

\section{Library Design}
\label{sec:library_design}

\placeholder builds on four core principles.

\paragraph{1. Built-in Post-hoc Interpretability} We include four state-of-the-art post-hoc feature importance methods and three interpretability corpora. Ready-to-use methods allow users to explain any text with an arbitrary model. Annotated datasets provide valuable test cases for new interpretability methods and metrics. To the best of our knowledge, \placeholder is first in providing integrated access to XAI datasets, methods, and a full-fledged evaluation suite. 

\paragraph{2. Unified Explanation Benchmarking} We propose a unified API to evaluate explanations. We currently support six state-of-the metrics along the principles of faithfulness and plausibility \citep{jacovi-goldberg-2020-towards}.

\paragraph{3. Transformers-readiness} \placeholder offers a direct interface with models from the Hugging Face Hub. Users can load models using standard naming conventions and explain them with the built-in methods effortlessly. Figure~\ref{fig:example_code} shows the essential code to classify and explain a string with a pre-existing Hugging Face model and evaluate the resulting explanations.

\paragraph{4. Modularity and Abstraction}
\placeholder counts three core modules, implementing Explainers, Evaluation, and Datasets APIs. Each module exposes an abstract interface to foster new development. For example, user can sub-class \texttt{BaseExplainer} or \texttt{BaseEvaluator} to include a new feature importance method or a new evaluation metric respectively. \\

\begin{table}[!t]
\centering
\begin{tabular}{p{0.52\linewidth}p{0.35\linewidth}}
\toprule
\textbf{Feature} & \textbf{Category} \\ \midrule
Gradient & Saliency \\ 
Integrated Gradient & Saliency \\ 
LIME & Surrogate Model \\
SHAP & Shapley Values \\ \midrule \midrule
Comprehensiveness  & Faithfulness \\
Sufficiency & Faithfulness \\
Correlation with \\
Leave-One-Out scores  & Faithfulness  \\
Intersection-Over-Union & Plausibility \\ 
Area Under \\ Precision-Recall Curve & Plausibility \\
Token-level F1 score & Plausibility \\ \midrule \midrule
HateXplain & Hate Speech \\
MovieReviews & Sentiment \\
SST & Sentiment \\ 
Thermostat & Generic \\ \bottomrule
\end{tabular}
\caption{\placeholder at a glance: built-in methods (top), metrics (middle), and datasets (bottom).}
\label{tab:built-in}
\end{table}

\placeholder builds on common choices from the interpretability community and good engineering practices. We report the most salient technical details (e.g., efficiency via GPU inference, visualization tools, etc.) in Appendix~\ref{app:technical_details}.  

\subsection{Explainer API}


We focus on the widely adopted family of post-hoc feature attribution methods \cite{danilevsky-etal-2020-survey}. I.e., given a model, a target class, and a prediction, \placeholder lets you measure \textit{how much} each token contributed to that prediction.
We integrate Gradient~\cite{simonyan2013deep} (also known as Saliency) and Integrated Gradient~\cite{sundararajan2017axiomatic};
SHAP~\cite{NIPS2017_lundberg_SHAP} as a Shapley value-based method, and LIME \cite{ribeiro2016should} as representative of local surrogate methods.

We build on open-source libraries and streamline their interaction with Hugging Face models and paradigms. We report the supported configurations and functionalities in Appendix~\ref{app:technical_details}.





\subsection{Dataset API}

Fostering a streamlined, accessible evaluation on independently released XAI datasets, we provide a convenient Dataset API.
It enables users to load XAI datasets, explain individual or subsets of samples, and evaluate the resulting explanations.

Currently, \placeholder includes three classification-oriented datasets annotated with human rationales, i.e., annotations highlighting the most relevant words, phrases, or sentences a human annotator attributed to a given class label \cite{deyoung-etal-2020-eraser,wiegreffe2021teach}. Moreover, \placeholder API gives access to the Thermostat collection \cite{feldhus-etal-2021-thermostat}, a wide set of pre-computed feature attribution scores.

\paragraph{HateXplain} \cite{mathew2021hatexplain}. It contains 20,148 English instances labeled along three axes: (i) hate (either hateful, offensive, normal or undecided), (ii) target group (either race, religion, gender, sexual orientation, or miscellaneous), and (iii) word-level human rationales (expressed only on hateful and offensive texts).\footnote{If a model splits a relevant word into sub-words, we consider all of them relevant as well.} 


\paragraph{MovieReviews} \cite{zaidan2008modeling, deyoung-etal-2020-eraser}. The dataset contains 2,000 movie reviews annotated with positive and negative sentiment labels and phrase-level human rationales that support gold labels.

\paragraph{Stanford Sentiment Treebank (SST)} \cite{socher2013recursive}. A sentiment classification dataset of 9,620 movie reviews annotated with binary sentiment labels, including human annotations for word phrases of the parse trees. We extract human rationales from annotations following the heuristic approach proposed in \newcite{carton-etal-2020-evaluating}.

\paragraph{Thermostat Datasets}
\textit{Thermostat} \cite{feldhus-etal-2021-thermostat} provides pre-computed feature attribution scores given a model, a dataset, and an explanation method. \placeholder currently provides built-in access to pre-computed attributions on the news topic classification and sentiment analysis tasks. \\

These datasets provide an \textit{initial} example of what an integrated approach can offer to researchers and practitioners.


\subsection{Evaluation API}
\label{ssec:eval_API}

We evaluate explanations on the faithfulness and plausibility properties \citep{jacovi-goldberg-2020-towards,deyoung-etal-2020-eraser}. Specifically, \placeholder implements three state-of-the-art metrics to measure faithfulness and three for plausibility. 

\paragraph{Faithfulness.}

Faithfulness evaluates how accurately the explanation reflects the inner working of the model \citep{jacovi-goldberg-2020-towards}.

\placeholder offers the following measures of faithfulness: comprehensiveness, sufficiency, \citep{deyoung-etal-2020-eraser} and correlations with ‘leave-one-out’
scores \citep{jain-wallace-2019-attention}.

\textit{Comprehensiveness} ($\uparrow$) 
evaluates whether the explanation captures the tokens the model used to make the prediction. We measure it by removing the tokens highlighted by the explainer and observing the change in probability as follows.

Let $x$ be a sentence and let $f_j$ be the prediction probability of the model $f$ for a target class $j$. 
Let $r_j$ be a discrete explanation or \textit{rationale} indicating the set of tokens supporting the prediction $f_j$. 
Comprehensiveness is defined as $f(x)_j - f(x \setminus r_j)_j$
where $x \setminus r_j$ is the sentence $x$ were tokens in $r_j$ are removed.
A high value of comprehensiveness indicates that the tokens in $r_j$ are relevant for the prediction.

While comprehensiveness is defined for discrete explanations, feature attribution methods assign a continuous score to each token. We hence select identify $r_j$ as follows. First, we filter out tokens with a negative contribution (i.e., they \textit{pull} the prediction away from the chosen label). Then, we compute the metric multiple times, considering the $k\%$ most important tokens, with $k$ ranging from 10\% to 100\% (step of 10\%).  
Finally, we aggregate the comprehensiveness scores with the average, called Area Over the Perturbation Curve (AOPC) \citep{deyoung-etal-2020-eraser}.

\textit{Sufficiency} ($\downarrow$) captures if the tokens in the explanation are sufficient for the model to make the prediction \citep{deyoung-etal-2020-eraser}.
It is measured as $f(x)_j - f(r_j)_j$. 
A low score indicates that tokens in $r_j$ are indeed the ones driving the prediction. As for \textit{Comprehensiveness}, we compute the AOPC by varying the number of the relevant tokens $r_j$.

\textit{Correlation with Leave-One-Out scores} ($\uparrow$). We first compute leave-one-out (LOO) scores by omitting tokens and measuring the difference in the model prediction. We do that for every token, once at a time.
LOO scores represent a simple measure of individual feature importance under the linearity assumption \citep{jacovi-goldberg-2020-towards}.
We then measure the Kendall rank correlation coefficient $\tau$ between the explanation and LOO importance \citep{jain-wallace-2019-attention} ($taucorr\_loo$). $taucorr\_loo$ closer to 1 means higher faithfulness to LOO.

\paragraph{Plausibility.}
Plausibility reflects how explanations are aligned with human reasoning by comparing explanations with \textit{human rationales}   \citep{deyoung-etal-2020-eraser} .

We integrate into \placeholder three plausibility measures of the ERASER benchmark \citep{deyoung-etal-2020-eraser}: Intersection-Over-Union (IOU) at the token level, token-level F1 scores, and Area Under the Precision-Recall curve (AUPRC).

The first two are defined for discrete explanations.
Given the human and predicted rationale, \textit{IOU} ($\uparrow$) quantifies the overlap of the tokens they cover divided by the size of their union.
\textit{Token-level F1 scores} ($\uparrow$) are derived by computing precision and recall at the token level.
Following \newcite{deyoung-etal-2020-eraser} and \newcite{mathew2021hatexplain}, we derive discrete explanations by selecting the top $K$ tokens with positive influence, where $K$ is the average length of the human rationale for the dataset.
While being intuitive, IOU and Token-level F1 are based only on a single threshold to derive rationales. Moreover, they do not consider tokens' relative ranking and degree of importance. 
We then also integrate the AUPRC ($\uparrow$), defined for  explanations with continuous scores \citep{deyoung-etal-2020-eraser}. It is computed by varying a threshold over token importance scores, using the human rationale as ground truth.

\subsection{Transformers-Ready Interface}

\placeholder is deeply integrated with Hugging Face interfaces. Users working with their standard models and tokenizers can easily integrate it for diagnostic purposes. The contact point is the main \texttt{Benchmark} class. It receives any Hugging Face model and tokenizer and uses them to classify, run explanation methods and seamlessly evaluate the explanations.
Similarly, our Dataset API leverages Hugging Face's \textit{datasets}\footnote{\url{https://github.com/huggingface/datasets}} to retrieve data and human rationales.

\lstdefinestyle{great}{
    backgroundcolor=\color{backcolour},   
    commentstyle=\color{codegreen},
    keywordstyle=\color{keywords},
    stringstyle=\color{nightblue},
    basicstyle=\ttfamily\footnotesize,
    breakatwhitespace=false,         
    breaklines=true,                 
    captionpos=b,                    
    keepspaces=true,                 
    showspaces=false,                
    showstringspaces=false,
    showtabs=false,                  
    tabsize=2,
    frame=shadowbox,
    emph={AutoTokenizer,AutoModelForSequenceClassification,Explainer},
    emphstyle={\color{emph}},
    emph={[2]from_pretrained,score,explain,show_table,evaluate_explanations,show_evaluation_table},
    emphstyle={[2]\color{lightpurple}},
    linewidth=7.5cm
}

\lstset{style=great}

\begin{figure}[!t]
\centering

\begin{lstlisting}[language=Python]
from transformers import AutoModelForSequenceClassification, AutoTokenizer
from ferret import Benchmark

name = "cardiffnlp/twitter-xlm-roberta-base-sentiment"
model = AutoModelForSequenceClassification.
    from_pretrained(name)
tokenizer = AutoTokenizer.
    from_pretrained(name)
    
bench = Benchmark(model, tokenizer)
query = "Great movie for a great nap!"

scores = bench.score(query)
print(scores)

# Run built-in explainers
explanations = bench.explain(
    query,
    target=2 # "Positive" label
)
bench.show_table(explanations)

# Evaluate explanations
evaluations = bench.evaluate_explanations(
    explanations, target=2
)
bench.show_evaluation_table(evaluations)

## Output
>> {'Negative': 0.013735532760620117,
>> 'Neutral': 0.06385018676519394,
>> 'Positive': 0.9224143028259277}
\end{lstlisting}


\centering
\includegraphics[frame,width=0.98\linewidth]{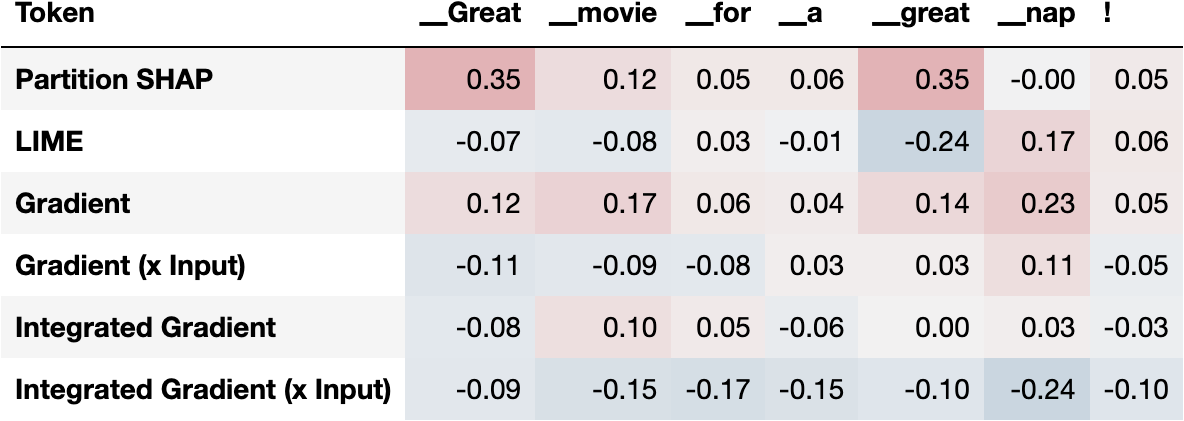}

\vspace{0.2cm}

\includegraphics[width=0.98\linewidth,frame]{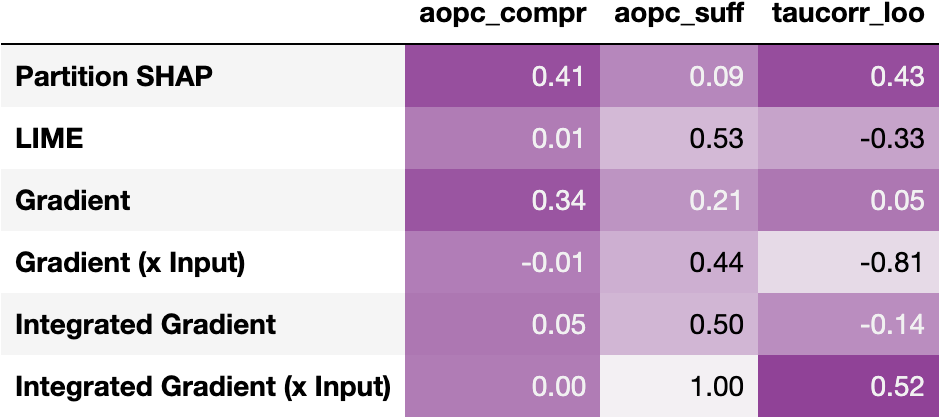}

\caption{Code to explain and evaluate explanations on a sentiment classifier (top). Token attributions (middle): darker red (blue) show higher (lower) contribution to the prediction. Faithfulness metrics (bottom): darker colors show better performance.}

\label{fig:code_instance}
\end{figure}

\section{Case Studies}

We showcase \placeholder in two real-world tasks, focusing on benchmarking explainers on individual samples or across multiple instances. In the following, we describe how \placeholder highlights the best explainers in sentiment analysis and hate speech detection tasks. Our running examples use an XLM-RoBERTa model fine-tuned for sentiment analysis \cite{DBLP:journals/corr/abs-2104-12250} and a BERT model fine-tuned for hate speech detection \cite{mathew2021hatexplain}. 

\subsection{Faithfulness Metrics for Error Analysis}

Explanations on individual instances are often used for model debugging and error analysis \cite{vig2019visualizing, feng-etal-2018-pathologies}. However, different explanations can lead users to different conclusions, hindering a solid understanding of the model's flaws. We show how practitioners can alleviate this issue including \placeholder in their pipeline.

Figure~\ref{fig:code_instance} shows explanations and faithfulness metrics computed on the sentence ``Great movie for a great nap!'' for the ``Positive'' class label misclassified by the model as ``Negative''.

Faithfulness metrics show that SHAP adheres best to the model's inner workings since it returns the most comprehensive and relevant explanations. Indeed, SHAP retrieves the highest number of tokens the model used to make the prediction ($aopc\_compr (\uparrow) =0.41$) that are relevant to drive the prediction ($aopc\_suff (\downarrow) =0.09$). Further, $taucorr\_loo (\uparrow) = 0.43$ indicates that SHAP explanations capture the most important tokens for the prediction under the linearity assumption. Although Integrated Gradient (x Input) shows a higher $taucorr\_loo$, it does not provide comprehensive and sufficient explanations. Similarly, Gradient and Integrated Gradient show bad sufficiency and comprehensiveness, respectively. LIME and Gradient (x Input) do not return trustworthy explanations according to all faithfulness metrics. 

Once SHAP has been identified as the best explainer, its explanations enable researchers to investigate possible recurring patterns or detect model biases thoroughly. In this case, the explanations shed light on a type of lexical overfitting: the word ``great'' skews the prediction toward the positive label regardless of the context and semantics. 

\subsection{Multi-Instance Assessment}

Instance-level analysis finds explainers that meet specific requirements locally. However, the best local explainer might be unsatisfactory across multiple instances.  
With \placeholder, users can easily produce and aggregate evaluation metrics across multiple dataset samples---or the entire corpus.

We describe how to choose the explainer that returns the most plausible and faithful explanations for the HateXplain dataset.
For demonstration purposes, we focus only on a sample of the dataset.
%

Figure~\ref{fig:ex_dataset} (Appendix~\ref{app:sec:results}) shows the metrics averaged across ten samples with the ``hate speech'' label.
Results suggest again that SHAP yields the most faithful explanations.
SHAP and Gradient achieve the best comprehensiveness and sufficiency scores, but SHAP outperforms all explainers for the $\tau$ correlation with LOO (\textit{taucorr\_loo} ($\uparrow$) = 0.41).
Gradient provides the most plausible explanations, followed by SHAP.

\begin{table*}[!t]
\centering
\begin{tabular}{@{}lccccc@{}}
\toprule
\textbf{} & \textbf{\thead{Multiple \\XAI approaches}} & \textbf{\thead{Transformers-\\readiness}} & \textbf{\thead{Evaluation \\APIs}} & \textbf{\thead{XAI \\datasets}} & \textbf{\thead{Built-in \\visualization}} \\ \midrule
Captum & \greencheck & \redcross & \redcross & \redcross & \greencheck \\
AllenNLP Interpret & \greencheck & \redcross & \redcross & \redcross & \redcross \\
Transformers-Interpret & \redcross & \greencheck & \redcross & \redcross & \greencheck \\
Thermostat & \greencheck & \greencheck & \redcross & \redcross & \greencheck \\ 
ContrXT & \redcross & \redcross & \redcross & \redcross & \redcross \\ 
OpenXAI & \greencheck & \redcross & \greencheck & \redcross & \redcross \\
NLPVis & \redcross & \redcross & \redcross & \redcross & \greencheck \\
Seq2Seq-Vis & \redcross & \redcross & \redcross & \redcross & \greencheck \\
BertViz & \redcross & \greencheck & \redcross & \redcross & \redcross \\
ELI5 & \redcross & \redcross & \redcross & \redcross & \greencheck \\
LIT & \greencheck & \redcross & \redcross & \redcross & \greencheck \\
ERASER & \redcross & \redcross & \greencheck & \greencheck & \redcross \\
Inseq & \greencheck & \greencheck & \redcross & \redcross & \greencheck \\
\textbf{\placeholder} & \greencheck & \greencheck & \greencheck & \greencheck & \greencheck \\ \bottomrule
\end{tabular}
\caption{Comparing off-the-shelf features across different XAI libraries. When assessing built-in visualization, we disregard tools that either do not provide a unified interface or provide single data-point visualizations.}
\label{tab:comparison}
\end{table*}

\section{Related Work}
\label{sec:related_work}

This section provides a review of tools and libraries that offer a subset of the \placeholder's functionalities, namely the option to use multiple XAI methods and datasets, evaluation API, transformer-readiness, and built-in visualization.
Table~\ref{tab:comparison} summarizes them and compares \placeholder with similar frameworks.



\paragraph{Tools for Post-Hoc XAI.}
Toolkits for post-hoc interpretability
offer built-in methods to explain model prediction, typically through a code interface. 
\placeholder builds on and extends this idea to a unified framework to generate explanations, \textit{evaluate and compare} them, with support to several XAI datasets. Moreover, \placeholder's explainers are integrated with transformers's \cite{wolf-etal-2020-transformers} principles and conventions.


PyTorch's Captum \citep{kokhlikyan2020captum} is a generic Python library supporting many interpretability methods. However, the library lacks integration with the Hugging Face Hub and offers no evaluation procedures.
AllenNLP Interpret \cite{wallace-etal-2019-allennlp} provides interpretability methods based on gradients and adversarial attacks for AllenNLP models \cite{gardner-etal-2018-allennlp}. We borrow the modular and extensible design and extend it to a wider set of explainers.
Transformers-Interpret\footnote{\url{https://github.com/cdpierse/transformers-interpret}} leverages Captum to explain Transformer models, but it supports only a limited number of methods.
Thermostat \cite{feldhus-etal-2021-thermostat} exposes pre-computed feature attribution scores through the Hugging-Face Hub but no features oriented to implement or evaluate XAI. We support the Thermostat as a third-party add-on and let users test and benchmark pre-computed explanations. Unlike our study, Inseq \citep{sarti-etal-2023-inseq} focuses on post-hoc interpretability for sequence generation models. Although researchers can use the library to add interpretability evaluations to their models, the toolkit lacks built-in evaluation metrics. 

Other related approaches enable global (rather than local) explainability \cite{malandri2022contrastive}, or explanation interfaces for non-transformers models on non-NLP tasks \cite{agarwal2022openxai}.
Other approaches study model behavior at the subgroup level~\cite{wang-etal-2021-textflint, goel-etal-2021-robustness,  pastor-divexplorer-2021, pastor-divexplorer-demo-2021}, focusing more on model evaluation and robustness rather than  its interpretation.

\paragraph{Visualization.}

Most studies that develop visualization tools to investigate the relationships among the input, the model, and the output focus either on specific NLP models - NLPVis \cite{liu2018visual}, Seq2Seq-Vis \cite{strobelt2018s}, or explainers - BertViz \cite{vig2019visualizing}, ELI5\footnote{\url{https://github.com/TeamHG-Memex/eli5}}. LIT \cite{tenney-etal-2020-language} streamlines exploration and analysis in different models. However, it acts mainly as a graphical browser interface. 
\placeholder provides a Python interface easy to integrate with pre-existing pipelines.

\paragraph{Evaluation.} 
Although prior works introduced diagnostic properties for XAI techniques, evaluating them in practice remains challenging. Studies either concentrate on specific model architectures \cite{lertvittayakumjorn-toni-2019-human, arras-etal-2019-evaluating, deyoung-etal-2020-eraser}, individual datasets \cite{guan2019towards, arras-etal-2019-evaluating}, or a single group of explainability methods \cite{robnik2018perturbation, adebayo2018sanity}. Hence, providing a generally applicable and automated tool for choosing the most suitable method is crucial. 
To this end, \citet{atanasova-etal-2020-diagnostic} present a comparative study of XAI techniques in three application tasks and model architectures. To the best of our knowledge, we are the first to present a user-friendly Python interface to interpret, visualize and empirically evaluate models directly from the Hugging Face Hub across several metrics.
We extend previous work from \citet{deyoung-etal-2020-eraser}, who developed a benchmark for evaluating rationales on NLP models called ERASER by offering a unified interface for evaluation \textit{and} visual comparison of the explanations at the instance- and dataset-level.

Closer to \placeholder, the OpenXAI framework~\cite{agarwal2022openxai} enables a systematic evaluation of feature attribution explanation, integrating multiple explainers and XAI structured datasets.
OpenXAI supports tabular datasets while we focus on textual data and NLP models.

\section{Conclusions}

We introduced \placeholder, a novel Python framework to easily access XAI techniques on transformer models.
With \placeholder, users can \textit{explain} using state-of-the-art post-hoc explainability techniques, \textit{evaluate} explanations on several metrics for faithfulness and plausibility, and easily \textit{interact} with datasets annotated with human rationales.

We built \placeholder with modularity and abstraction in mind to facilitate future extensions and contributions from the community (see Appendix \ref{app:sec:ongoing} for an overview of the ongoing development). As future work, we envision off-the-shelf support for new NLP tasks and scenarios.   
Building on the classification setup presented in this paper, we plan to add support to more NLP tasks that can be framed as classification, such as Mask Filling Prediction, Natural Language Inference, Zero-Shot Text Classification, Next Sentence Prediction, Token Classification, and Multiple-Choice QA. 
One further direction would be improving \placeholder's interoperability with new libraries, e.g., Inseq \cite{sarti-etal-2023-inseq} for XAI on text generation tasks and models.   


\section*{Ethics Statement}

\placeholder's primary goal is to facilitate the comparison of methods that are instead frequently tested in isolation. Nonetheless, we cannot assume the metrics we currently implement provide a full, exhaustive picture, and we work towards enlarging this set accordingly.

Further, interpretability is much broader than post-hoc feature attribution. We focus on this family of approaches for their wide adoption and intuitiveness.  


Similarly, the evaluation measures we integrate are based on removal-based criteria. Prior works pointed out their limitations, specifically the problem of erased inputs falling out of the model input distribution \citep{NEURIPS2019_fe4b8556}.

\section*{Acknowledgments}

This project has partially received funding from the European Research Council (ERC) under the European Union’s Horizon 2020 research and innovation program (grant agreement No.\ 949944, INTEGRATOR), by Fondazione Cariplo (grant No. 2020-4288, MONICA), and by the grant ``National Centre for HPC, Big Data and Quantum Computing'', CN000013 (approved under the M42C Call for Proposals - Investment 1.4 - Notice ``National Centers'' - D.D. No. 3138, 16.12.2021, admitted for funding by MUR Decree No. 1031, 17.06.2022). 
DN and GA are members of the MilaNLP group and the Data and Marketing Insights Unit of the Bocconi Institute for Data Science and Analysis.
EP did part of the work while at CENTAI and is currently a member of the DataBase and Data Mining Group (DBDMG) at Politecnico di Torino. CDB contributed to the work while at Bocconi University and is currently part of the UKRI Centre for Doctoral Training in Safe and Trusted Artificial Intelligence (www.safeandtrustedai.org). 

\bibliography{anthology,custom}
\bibliographystyle{acl_natbib}

\appendix

\section{Technical Details}
\label{app:technical_details}

\subsection{Explainer API}
\label{app:sec:explainer}

Our implementation is built on top of original implementations (as for SHAP and LIME) and open-source libraries (as Captum \citep{kokhlikyan2020captum} for gradient-based explainers) to directly explain Transformer-based language models.

Currently, we integrate Gradient (G)~\cite{simonyan2013deep}, Integrated Gradient (IG)~\cite{sundararajan2017axiomatic}, SHAP~\cite{NIPS2017_lundberg_SHAP}, and LIME \cite{ribeiro2016should}. For G and IG, users can get explanations from plain gradients or multiply gradients by the input token embeddings. For SHAP, we use the Partition approximation to estimate Shapley values.\footnote{\url{https://shap.readthedocs.io/en/latest/generated/shap.explainers.Partition.html}}

\subsubsection{Evaluation API}

While human gold annotations are normally discrete, current explainers provide continuous token attribution scores. Following previous work, we hence go from continuous scores to a discrete set of \textit{relevant} tokens (i.e., $r_j$ in Section~\ref{ssec:eval_API}) as follows.   

We consider only tokens with a positive contribution to the chosen label (i.e., they \textit{push} the prediction towards the chosen label).
For the AOPC comprehensiveness and sufficiency measures, the relevant tokens in the discrete rationale are the most $k\%$ important tokens with $k$ ranging from 10\% to 100\% (step of 10\%). 
For token-level IOU and F1 scores plausibility measure, we follow the  \newcite{deyoung-etal-2020-eraser} and \newcite{mathew2021hatexplain} approach, and we select the top $k$ tokens where $k$ is the average length of human rationales for the dataset. 

The evaluation measures at the dataset level are the average scores across explanations. 
Differently than \newcite{deyoung-etal-2020-eraser} that use the F1 IOU score, we directly compute the average token-level IOU.

All human rationales are at the token level, indicating the most relevant tokens to a given class label.

\begin{figure*}[!t]
    \centering
    \includegraphics[width=0.8\linewidth,frame]{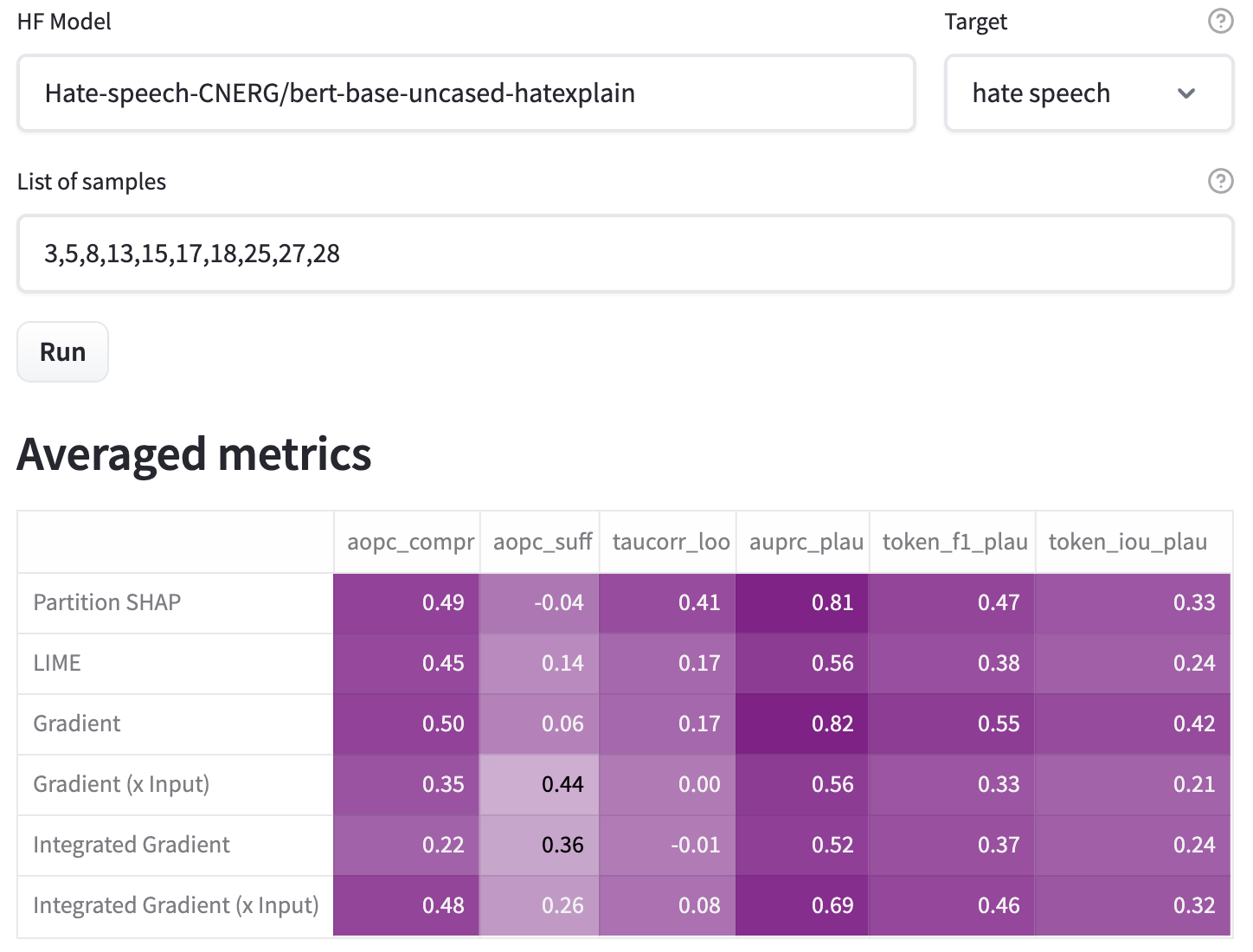}
    \caption{Faithfulness and Plausibility metrics averaged across ten samples with the ``hateful'' label of HateXPlain. Darker colors mean better performance.}
    \label{fig:ex_dataset}
\end{figure*}

\subsection{Technical Features}

\placeholder implements several functionalities to facilitate end users in using it.

\begin{itemize}
    \item High-level interface. Most of \placeholder's features, such as interpretability methods and evaluation measures, are accessible via a single entry point, the \texttt{Benchmark} class.
    \item GPU-enabled batched inference. \placeholder requires running inference for certain executions. It uses batching and local GPUs transparently to the user whenever that happens.
    \item Visualization methods. The \texttt{Benchmark} class exposes several methods to visualize attribution scores and evaluation results in tabular format. These tables are plotted seamlessly on Jupyter Notebooks (see Figure~\ref{fig:code_instance} (bottom) for an example).  
    
\end{itemize}

\section{Ongoing Development}
\label{app:sec:ongoing}

\placeholder is under active development. We are extending the  core modules as follows.

\paragraph{Explainers.}
We plan to integrate two recent interpretability methods that require training a complementary model. Sampling and Occlusion (SOC) \cite{jin2019towards} provides a hierarchical explanation to address compositional contributions. 
Minimal Contrastive Editing (MiCE) \cite{ross-etal-2021-explaining} trains a T5 \cite{raffel2020exploring} model to implement contrastive edits to the input to change the model output.
Finally, we are including a third gradient-based algorithm. Integrated Discretized Gradients \cite{sanyal-ren-2021-discretized} improve IG sampling intermediate steps close to actual words in the embedding space.

\paragraph{Evaluators.}

We plan to include additional evaluation measures such as sensitivity, stability \cite{yin-etal-2022-sensitivity}, and Area Under the Threshold-Performance curve (AUC-TP) \cite{atanasova-etal-2020-diagnostic}. 

\section{Additional Results}
\label{app:sec:results}

Figure~\ref{fig:ex_dataset} shows a screenshot of dataset-level assessment from our demo web app. It reports the evaluation metrics averaged across ten samples with the ``hate speech'' label for the HateXplain dataset, discussed in Section \ref{fig:ex_dataset}.

The user specifies a model from the Hugging Face Hub (\textit{HF Model} field), the target class (\textit{Target}), and the set of samples of interest (\textit{List of samples}). 
\placeholder web app directly computes explanation and their evaluation and visualizes the results.




\end{document}